\documentclass{midl}

\usepackage{framed,multirow}

\usepackage{amssymb}
\usepackage{latexsym}

\usepackage{url}

\RequirePackage{amsmath}
\usepackage{makeidx}  
\usepackage{multirow}

\usepackage{hhline}
\usepackage[font=small]{caption}
\usepackage{subcaption}
\usepackage{graphicx}
\usepackage{floatrow}

\usepackage{mwe} 

\begin{document}

\midlauthor{\Name{Seyed Raein Hashemi\nametag{$^{1,2}$}},
\Name{Sanjay P. Prabhu\nametag{$^{1}$}},
\Name{Simon K. Warfield\nametag{$^{1}$}},
\Name{Ali Gholipour\nametag{$^{1}$}}\\
\addr $^{1}$ Computational Radiology Laboratory, Boston Children's Hospital; and Harvard Medical School \\
\addr $^{2}$ College of Computer and Information Science, Northeastern University; Boston, MA
}

\title[Exclusive Independent Probability Estimation: IsoIntense Infant Brain Segmentation]{Exclusive Independent Probability Estimation using Deep 3D Fully Convolutional DenseNets: Application to IsoIntense Infant Brain MRI Segmentation}

\maketitle

\begin{abstract}
The most recent fast and accurate image segmentation methods are built upon fully convolutional deep neural networks. In particular, densely connected convolutional neural networks (DenseNets) have shown excellent performance in detection and segmentation tasks. In this paper, we propose new deep learning strategies for DenseNets to improve segmenting images with subtle differences in intensity values and features. In particular, we aim to segment brain tissue on infant brain MRI at about 6 months of age where white matter and gray matter of the developing brain show similar T1 and T2 relaxation times, thus appear to have similar intensity values on both T1- and T2-weighted MRI scans. Brain tissue segmentation at this age is, therefore, very challenging. 
To this end, we propose an exclusive multi-label training strategy to segment the mutually exclusive brain tissues with similarity loss functions that automatically balance the training based on class prevalence. Using our proposed training strategy based on similarity loss functions and patch prediction fusion we decrease the number of parameters in the network, reduce the complexity of the training process focusing the attention on less number of tasks, while mitigating the effects of data imbalance between labels and inaccuracies near patch borders.
By taking advantage of these strategies we were able to perform fast image segmentation (less than 90 seconds per 3D volume), using a network with less parameters than many state-of-the-art networks (1.4 million parameters), overcoming issues such as 3D vs 2D training and large vs small patch size selection, while achieving the top performance in segmenting brain tissue among all methods tested in first and second round submissions of the isointense infant brain MRI segmentation (iSeg) challenge according to the official challenge test results.
Our proposed strategy improves the training process through balanced training and by reducing its complexity while providing a trained model that works for any size input image, and is fast and more accurate than many state-of-the-art methods. 

\end{abstract}

\begin{keywords}
Deep learning, Convolutional Neural Network, FC-DenseNet, Segmentation
\end{keywords}


\section{Introduction}
\label{sec:intro}
Deep convolutional neural networks have shown great potential in medical imaging on account of dominance over traditional methods in applications such as segmentation of neuroanatomy \citep{bui2017dense,moeskops2016automatic,zhang2015deep,chen2017voxresnet}, lesions \citep{valverde2017improving,brosch2015deep,kamnitsas2017efficient,hashemi2018asym}, and tumors \citep{havaei2017brain,pereira2016brain,wachinger2017deepnat} using voxelwise networks \citep{moeskops2016automatic,havaei2017brain,salehi2017auto,salehi2018real}, 3D voxelwise networks \citep{chen2017voxresnet,kamnitsas2017efficient} and Fully Convolutional Networks (FCNs) \citep{cciccek20163d,milletari2016v,salehi2017auto,salehi2018real,hashemi2018asym}. FCNs have shown better performance while also being faster in training and testing than voxelwise methods~\citep{salehi2017auto,salehi2018real}. 

Among these, the densely connected networks, referred to as  DenseNets~\citep{huang2017dense} and a few of its extensions, such as a 3D version called DenseSeg~\citep{bui2017dense} and a fully convolutional two-path edition (FC-DenseNet)~\citep{jegou2017tiramisu}, have shown promising results in image segmentation tasks~\cite{dolz2018hyper}. For example the DenseSeg showed top performance in the 2017 MICCAI isointense infant brain MRI segmentation (iSeg) grand challenge\protect\footnotemark[1], which is considered a very difficult image segmentation task for both traditional and deep learning approaches. During early infant brain maturation through the myelination process, there is an isointense period in which the T1 and T2 relaxation times of the white matter (WM) and gray matter (GM) tissue become similar, resulting in isointense (similar intensity) appearance of tissue on both T1-weighted and T2-weighted MRI contrasts. This happens around 6 months of age where tissue segmentation methods that are based directly on image intensity are prone to fail~\citep{Wang2013neo}. Deep learning methods, however, have shown promising results in this application. 
\footnotetext[1]{\url{http://iseg2017.web.unc.edu/}}
\footnotetext[2]{\url{http://iseg2017.web.unc.edu/rules/results}}
\footnotetext[3]{\url{http://iseg2017.web.unc.edu/evaluation-on-the-second-round-submission}} 

In this work, we aimed to further improve image segmentation under these challenging conditions. While the top performing methods in the iSeg challenge relied on the power of DenseNets and used conventional training strategies based on cross-entropy loss function~\cite{bui2017dense,dolz2018hyper}, in this work we focused on the training part and developed new strategies that helped us achieve the best performance currently reported on the iSeg challenge among all first\protect\footnotemark[2] and second round submissions\protect\footnotemark[3]. We built our technique over a deep 3D two-channel fully convolutional DenseNet; and trained it purposefully using our proposed exclusive multi-label multi-class method of training, with exclusively adjusted similarity loss functions on large overlapping 3D image patches. We overcame class similarity issues by focusing the training on one of the isointense class labels (WM) instead of both (thus referred to as exclusive multi-label multi-class), where we balanced precision and recall separately for each class using $F_\beta$ loss functions~\cite{hashemi2018asym} with $\beta$ values adjusted with respect to class prevalence in the training set.

Our contributions that led to improved iso-intense image segmentation include 1) An exclusive multi-label multi-class training approach (through independent probability estimation) using automatically-adjusted similarity loss functions per class; 2) utilizing a 3D FC-DenseNet architecture adopted from~\cite{jegou2017tiramisu} that is deeper, has more skip connections and has less parameters than networks in previous studies; and 3) training and testing on large overlapping 3D image patches with a patch prediction fusion strategy~\cite{hashemi2018asym} that enabled intrinsic data augmentation and improved segmentation in patch borders while having the advantage of using any size image. Similarity loss functions, such as the Dice similarity loss, were previously proposed for two-class segmentation in V-Net~\citep{milletari2016v}. The $F_\beta$ loss functions, which showed excellent performance in training deep networks for highly imbalanced medical image segmentation~\cite{hashemi2018asym}, appeared to be effective also in exclusive multi-label training of DenseNets for independent multi-class segmentation in this work, where the class imbalance hyper-parameter $\beta$ was directly adjusted based on training data in the training phase.



The official results on iSeg test data show that our method outperformed all previously published and reported methods improving DenseNets while standing in the first place after the second round submissions as of December 2018. Our proposed training strategy can be extended to other applications for independent multi-class segmentation and detection with multiple very similar and unbalanced classes. After a brief technical description of the isointense infant brain MRI segmentation challenge in the Motivation, the details of the methods, including the network architecture and our proposed strategies for training are presented in the Methods section; and are followed by the Results and Conclusions.

\section{Motivation}

The publicly available MICCAI grand challenge on 6-month infant brain MRI segmentation (iSeg) dataset contains pre-processed T1- and T2-weighted MR images of 10 infant subjects with manual labels of cerebrospinal fluid (CSF), WM, and GM for training and 13 infant subjects without labels for testing which are all pre-defined by challenge officials. The intensity distribution of all classes (CSF, GM, WM) is shown in Figure~\ref{fig:intensity} in Appendix~\ref{sec:tablesfigures}, which shows that the intensity values of GM and WM classes on both MRI scans largely overlap. The GM-WM isointense appearance only happens around this stage of brain maturation and hinders GM-WM segmentation. CSF, which shows less overlap with GM and WM, shows a relatively spread intensity distribution, which is partly attributed to partial voluming effects in relatively large voxels where signal is averaged in CSF-GM interface, and inclusion of some other tissues such as blood vessels in the CSF label in iSeg data.

In the iSeg training data, the number of voxels in each class label is different and can be roughly presented as the average ratio of $(36, 1, 2, 1.5)$ for non-brain, CSF, GM, and WM classes, respectively. Unbalanced labels can make the training process converge to local minima resulting in sub optimal performance. The predictions, thus, may lean towards the GM class especially when distinguishing between the isointense areas of GM and WM. Using our proposed exclusive multi-label multi-class training method, which can be extended to other segmentation or detection tasks with very similar (isointense) while exclusive labels (each voxel belonging to a single label), we aimed to 1) let the network focus on and learn one of the segmentation challenges at a time rather than two (in this case WM rather than both GM and WM), 
2) reduce the bias in training towards classes with higher prevalence (in this case GM), and 3) use dedicated impartial asymmetric similarity loss functions on each of the non-similar classes independently (in this case WM and CSF).

\section{Methods}

\subsection{Network architecture}
In traditional densely connected networks each layer is connected to every other layer to preserve both high- and low-level features, in addition to allowing the gradients to flow from bottom layers to top layers resulting in more accurate predictions. Unlike Resnets~\cite{he2018resnet} which only sum the output of the identity function at each layer with a skip connection from the previous layer, DenseNets~\cite{huang2017dense} significantly improve the flow of information throughout the network by 1) using concatenation instead of summation and 2) forward connections from all preceding layers rather than just a previous layer, therefore:
\begin{equation}
    x_l^{(Resnet)} = H_l(x_{l-1}) + x_{l-1} \;\;\;\;\;,\;\;\;\;\; x_l^{(Densenet)} = H_l([x_0, x_1, ..., x_{l-1}])
\end{equation}
\noindent where $x_l$ is the output of the $l^{th}$ layer, $H_l$ is the $l^{th}$ layer transition, and $[x_0, x_1, ..., x_{l-1}]$ refers to the concatenation of all previous feature maps.


\begin{figure}
    \centering
    \includegraphics[width=0.95\textwidth]{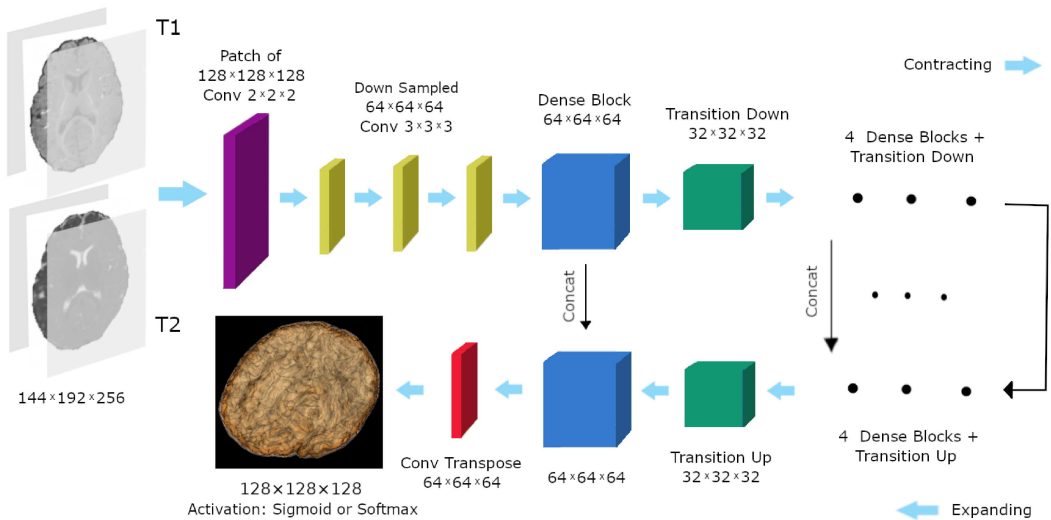}
    \caption{The 3D FC-DenseNet architecture used in this study; In the first layer, the input patch is downsampled from $128 \times 128 \times 128$ to $64 \times 64 \times 64$ using a $2 \times 2 \times 2$ convolution with stride 2 (purple). Before the activation layer the patch is upsampled from $64 \times 64 \times 64$ to $128 \times 128 \times 128$ using a $2 \times 2 \times 2$ convolution transpose with stride 2 (red). Using this deep architecture, we mitigated memory size limitations with large input patches, while maintaining a large field of view and incorporating 5 skip connections to improve the flow of local and global feature information.}
    \label{fig:Net}
\end{figure}

We designed our deep 3D densely connected network based on a combination of DenseSeg~\cite{bui2017dense} and FC-DenseNet~\cite{jegou2017tiramisu} architectures. This deep DenseNet~\cite{huang2017dense} style architecture is shown in Figure~\ref{fig:Net} consisting of contracting and expanding paths. 
The network is trained on local features in the contracting path concatenated with upsampled global features in the expanding path. For this reason, the model has the capacity to learn both high-resolution local and low-resolution global 3D features. The depth of the architecture as well as the 5 skip-layer connections ensure the use of 5 various resolutions of local and global 3D features in the final prediction.


The contracting path contains an initial $2\times2\times2$ convolution with stride 2 for the purpose of downsampling and reducing the patch size (to $64 \times 64 \times 64$) while preserving the larger field of view ($128 \times 128 \times 128$). It is then followed by three padded $3\times3\times3$ convolutional layers. Five dense blocks follow with a growth rate of 12. Growth rate for dense blocks is the increase amount in the number of feature maps after each block. Dense blocks contain four $3\times3\times3$ convolutional layers preceding with $1\times1\times1$ convolutions which are known as bottlenecks~\cite{huang2017dense}. Dimension reduction of 0.5 applied at transition layers helps parameter efficiency and reduce the number of input feature maps. There are skip connections between each and every layer inside dense blocks. Aside from the center dense block connecting the two paths, contracting dense blocks are followed by a $1\times1\times1$ convolutional layer and a max pooling layer referred to as transition down blocks, and expanding dense blocks are preceded with a $3\times3\times3$ transpose convolution with stride 2 known as transition up blocks~\cite{jegou2017tiramisu}. All convolutional layers in the network are followed by batch normalization and ReLU non-linear layers. Dropout rate of 0.2 is used only after $3\times3\times3$ convolutional layers of each dense block. As the final layer a $1\times1\times1$ convolution with a sigmoid or softmax output is used, which is discussed later.

\subsection{Training strategy}
While our deep two-channel 3D FC-DenseNet architecture used two extra downsampling and upsampling convolutional layers (purple and red layers in Figure~\ref{fig:Net}) to preserve higher fields of view with more skip connections than other DenseNet variants \cite{bui2017dense,dolz2018hyper}, in this work we mainly focused on innovative training strategies to facilitate network training and improve performance. These innovations constitute two training approaches, i.e. single-label multi-class and exclusive multi-label multi-class training, with asymmetric similarity loss functions based on $F_\beta$ scores~\cite{hashemi2018asym}, use of large image patches as input, and a patch prediction fusion strategy, which are discussed here.

\subsubsection{Single-label multi-class}
Often in machine learning and deep learning tasks, all labels in a dataset are mutually exclusive which is also the case for the iSeg dataset. This is called a single-label multi-class problem where each voxel can only have one label inside a multi-class environment. One of the most important decisions in a network is the choice of the classification layer. The usual choice for this type of classification for image segmentation is a softmax layer which is a normalized exponential function and a generalization of the logistic function forcing probability values to be in the range of [0,1] with the total class probability sum of 1. Softmax assumes independability of each class to other classes, which is theoretically accurate in the case of iSeg labels (CSF, GM, WM). However, because of human error in generating accurate ground truth labels as well as the isointense specification of GM and WM classes in 6-month infant MRIs, incorporating this theory could result in complications on the border voxels of the two labels where the intensities are most analogous.

In the single-label approach we trained the network the more popular way to learn all labels together with a softmax activation function as shown in Figure~\ref{fig:softmaxsigmoid}(a), where the highest probability class was selected for each voxel. Even though we used an asymmetric loss function to account for data imbalance (discussed later), the network applied the required precision-recall asymmetry mostly on labels with higher level of occurrence since all the labels were trained together. In this case the GM label being the most prevalent class (46.7\% of all labeled voxels), receives higher recall than the other labels (21.84\% CSF and 31.45\% WM prevalence). Considering both the level of occurrence as well as the isointense aspect of infant brain MRIs, the WM class would receive the least recall among all labels. Therefore, we aimed to exploit other strategies, in particular exclusive independent probability estimation using a multi-label multi-class strategy to better balance the training.

\subsubsection{Multi-label multi-class}
Unlike single-label problems where voxels can only have one label, in multi-label multi-class problems each voxel has the potential to have multiple labels in a multi-class environment. These types of tasks require prediction of multiple labels per voxel. By using softmax as the activation function, a constant threshold cannot be used practically because the probabilities are not evenly distributed for every patch or image. Therefore, some sort of binary classification or output function is needed; such as the sigmoid function:
\begin{equation}
    \sigma_z^{softmax} = \frac{e^z}{\Sigma_{k=1}^ne^k} \;\;\;\; , \;\;\;\; \sigma_z^{sigmoid} = \frac{1}{1 + e^{-z}} = \frac{e^z}{1 + e^z} = \frac{e^z}{e^0 + e^z}
\label{eq:sigmoid}
\end{equation}


\noindent where $\sigma$ denotes the output of the softmax and sigmoid functions, $z$ is the output for label z before activation, $k$ is the output for each label $\in [1,n]$ before activation and $n$ is the total number of labels. Sigmoid is a special case of softmax for only two classes (i.e. 0 and z) which models the probability of classes as Bernoulli distributions and independent from other class probabilities. In the multi-label approach, instead of training all the class labels 
to a probability sum of 1, we scale each class probability separately between [0,1] so we can use a constant threshold to extract labels. The multi-label multi-class approach has two other advantages: 1) different loss functions and hyper-parameters can be used for distinct training of classes; and 2) calculating sigmoid is less computationally cumbersome for a processing unit compared to softmax especially for large number of labels.

\subsubsection{Exclusive multi-label multi-class}

\begin{figure}
\begin{subfigure}
    \centering
    \includegraphics[width=.45\textwidth]{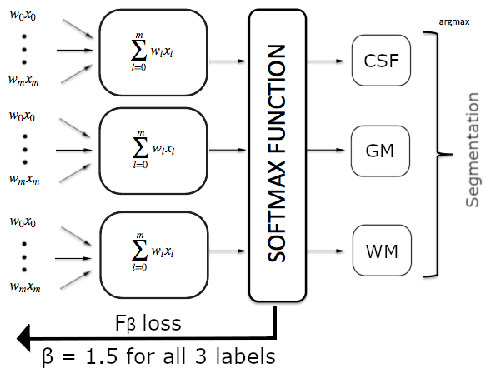}
\end{subfigure}%
\begin{subfigure}
    \centering
    \includegraphics[width=.45\textwidth]{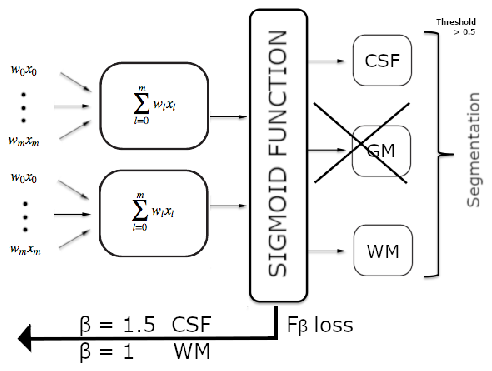}
\end{subfigure}
\caption{Examples of the a) Single-label approach with the Softmax activation function and single loss function for all labels, and b) Exclusive Multi-label approach with the Sigmoid activation function and multiple loss function configurations for different labels in iSeg segmentation task.}
\label{fig:softmaxsigmoid}
\end{figure}

Since we decided to use a less complex cost function and train the class labels independently, there was no reason to train on both of the isointense labels, especially as the classes were mutually exclusive. In fact reducing one of the classes helps the network focus its attention to one label while eliminating the effect of biased learning towards a class with higher prevalence. This way, the model has an easier task of learning subtle differences between nearly indistinguishable classes such as GM and WM in isointense infant brain MRI segmentation. This can potentially be generalizable to every combination of extremely hard to detect, unbalanced, and mutually exclusive class labels, excluding the one with more occurrences and training on the other while reducing both the number of network parameters and the complexity of the training process. To this end, for the iSeg data, as shown in Figure~\ref{fig:softmaxsigmoid}(b) we removed GM from the training cycle and trained the CSF and WM classes against non-CSF and non-WM labels using the Sigmoid activation function with differently balanced similarity loss functions discussed in the next section. The GM labels were concluded from the compliment of the already predicted CSF and WM labels.

\subsubsection{Loss function}
\label{sec:loss}
To better deal with data imbalance, we used an extension of the idea of using Dice similarity loss~\cite{milletari2016v}, based on the $F_\beta$ scores~\cite{hashemi2018asym} defined as:
\begin{equation}
    F_\beta = \frac{(1+\beta^2) \times precision \times recall}{\beta^2 \times precision+recall} =  \frac{TP}{TP + (\frac{\beta^2}{1+\beta^2}) FN + (\frac{1}{1+\beta^2}) FP}
\label{eq:fbeta}
\end{equation}
\noindent Where $TP$, $FN$, and $FP$ are true positives, false negatives, and false positives, respectively. While Dice similarity is a harmonic mean of precision and recall, $F_\beta$ allows balancing between precision and recall by an appropriate choice of the hyper-parameter $\beta$. The selection method for $\beta$ values based on class prevalence is explained in Appendix~\ref{sec:hyper}.

\subsubsection{3D large patches and patch prediction fusion}
Rather than training on full-size, two-channel (T1- and T2-weighted MRI) input images, we extracted and used large 3D two-channel image patches as inputs and augmented the training data at the level of large patches. This had several advantages including efficient use of memory, intrinsic data augmentation, and the design of an image size-independent model. Previously in the Network Architecture section, we mentioned that large patches of $128 \times 128 \times 128$ were selected from the image and were immediately downsampled through a convolutional layer within the network to the size of $64 \times 64 \times 64$ in order to preserve higher receptive field while adapting to GPU memory restrictions. Nonetheless, accuracy near patch borders was relatively low mainly because of the effective receptive field of patches. To circumvent this problem while fusing patch predictions in both training and testing, we exploited a weighted soft voting approach~\cite{hashemi2018asym} using second-order spline functions placed at the center of patches. Patches were selected for prediction using $50\%$ overlaps. Each patch was rotated 180 degrees once in each direction for augmentation in both training and testing, and the final result was calculated through the fusion of predictions by all overlapping patches and their augmentations (32 predictions per voxel).

\subsection{Experimental design}

We trained and tested our 3D FC-DenseNet with $F_\beta$ loss layer to segment isointense infant brains. T1- and T2-weighted MRI of 10 subjects were used as input, where we used five-fold cross-validation in training. There was not any pre-processing involved as the images were already skull-stripped and registered. The two MRI images of each subject were normalized through separately dividing each voxel value by the mean of non-zero voxels in each image. This way the whole brain (excluding background) in each modality was normalized to unit mean. Our 3D FC-DenseNet was trained end-to-end. Cost minimization on 2,500 epochs 
was performed using ADAM optimizer~\cite{kingma2014adam} with an initial learning rate of 0.0005 multiplied by 0.9 every 500 steps. The training time for this network was approximately 14 hours on a workstation with Nvidia Geforce GTX1080 GPU.

Validation and test volumes were segmented using feed-forward through the network. The output of the last convolutional layer with softmax non-linearity consisted of a probability map for CSF, GM and WM tissues. For the sigmoid version of the network, it contained only the CSF and WM tissues. In the softmax approach (single-label multi-class), the class with maximum probability among all classes was selected as the segmentation result for each voxel, while in the sigmoid approach (exclusive multi-label multi-class) voxels with computed probabilities $\geq0.5$ were considered to belong to the specific tissue class (CSF or WM) and those with probabilities $<0.5$ were considered non tissue. For voxels with both CSF and WM probabilities of $\geq0.5$ the class with higher probability was selected. Finally, GM labels were generated based on the compliment of predicted CSF and WM class labels. For evaluation, following iSeg, we report the Dice Similarity Coefficient (DSC), Hausdorff Distance (HD), and the Average Surface Distance (ASD) all defined in Appendix~\ref{sec:metrics}. 


\section{Results}
To evaluate the effect of our proposed exclusive multi-label multi-class training strategy compared to the single-label (without exclusive multi-label) method, we trained our FCN with single- and multi-label designs and calculated cross-validation DSC. The characteristics and performance metrics of our two trained models are compared in Appendix~\ref{sec:tablesfigures} Table~\ref{table:comp}, along with a comparison of other methods on a different validation set from~\cite{bui2017dense}. Paired sample t-test between exclusive multi-label and single-label configurations confirmed that differences were statistically significant ($p < 0.05$) among DSC, sensitivity and specificty results of CSF, GM and WM. The results in the top part of Table~\ref{table:comp} and our results in the bottom part should not be directly compared as they are on different validation sets. The actual comparison based on official iSeg test set are reported in Table~\ref{table:res}.

\begin{figure*}
    \centering
    \includegraphics[width=.7\textwidth]{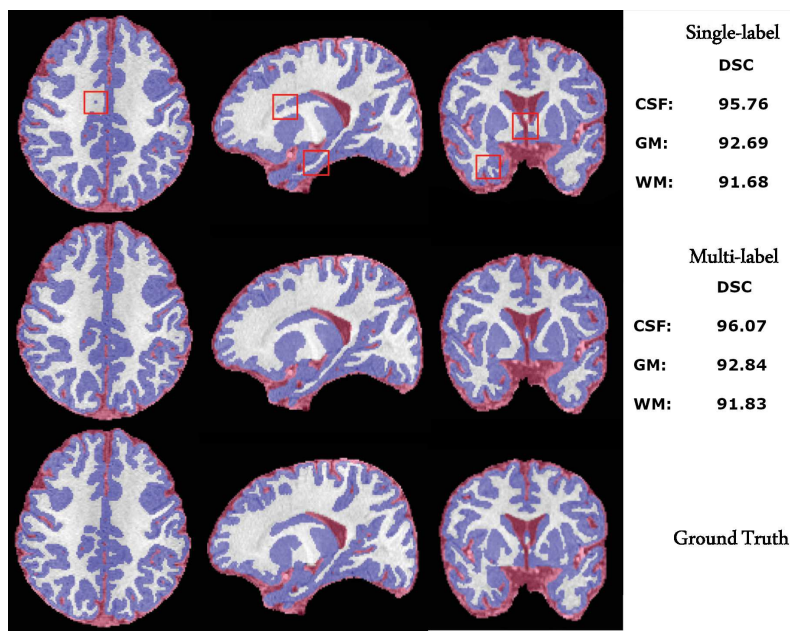}
    \caption{Segmentation results of 3D FC-DensNet of subject 1 in the validation set with Exclusive Multi-label and without Exclusive Multi-label (Single-label). Dice scores for each class label are also shown for each image. Red squares highlight some areas of differences between the two approaches.}
    \label{fig:res}
\end{figure*}

The official challenge test results on challenge website, reported in Table~\ref{table:res} for the top performing teams, show that our approach outperformed all first and second round submissions including the DenseNet based methods of DenseSeg~\cite{bui2017dense} and HyperDenseNet~\cite{dolz2018hyper}. According to the DSC and ASD performance metrics our exclusive multi-label method 
performed better than all other methods for all (CSF, GM and WM) classes. The results of the HD score, however, were not consistent; nonetheless the HD score
is not an appropriate performance measure for segmentation of complex shapes based on the comprehensive discussion and evaluation in~\cite{taha2015metrics} (Appendix~\ref{sec:metrics}). Overall, official challenge results show improved segmentation in iSeg using our method, which is attributed to 1) our 3D FC-DenseNet architecture which is deeper than previous DenseNets with more skip-layer connections and less number of parameters; and more importantly 2) our proposed exclusive multi-label training with $F_\beta$ loss functions that made a better balance between precision and recall in training the network. Figure~\ref{fig:res} shows prediction results of a subject from one of the validation folds for our two training methods compared to the ground truth. Visual assessment and the DSC scores on all labels consistently show that the best results were achieved by our exclusive multi-label model.

\floatsetup[table]{capposition=top}

\begin{table*}[!ht]
\small
\centering
 \begin{tabular}{|c||c|c|c|c|c|c|c|c|c|} 
 \hline
 & \multicolumn{3}{c|}{CSF} & \multicolumn{3}{c|}{GM} & \multicolumn{3}{c|}{WM}\\
 \hline
Teams (Published) & DSC & HD & ASD & DSC & HD & ASD & DSC & HD & ASD\\
 \hhline{|==========|}
BCH CRL Imagine (ours) & \textbf{96.0} & \textbf{8.85} & \textbf{0.11} & \textbf{92.6} & 9.55 & \textbf{0.31} & \textbf{90.7} & 7.1 & \textbf{0.36} \\
\hline
MSL SKKU (2nd rnd) & 95.8 & 9.11 & 0.116 & 92.3 & 6.0 & 0.32 & 90.4 & 6.62 & 0.375 \\
\hline
MSL SKKU (1st rnd) & 95.8 & 9.07 & 0.116 & 91.9 & 5.98 & 0.33 & 90.1 & \textbf{6.44} & 0.39 \\
\hline
LIVIA (2nd rnd) & 95.6 & 9.42 & 0.12 & 92.0 & \textbf{5.75} & 0.33 & 90.1 & 6.66 & 0.38 \\
\hline
LIVIA (1st rnd) & 95.7 & 9.03 & 0.138 & 91.9 & 6.42 & 0.34 & 89.7 & 6.98 & 0.38 \\
\hline
Bern IPMI & 95.4 & 9.62 & 0.127 & 91.6 & 6.45 & 0.34 & 89.6 & 6.78 & 0.4 \\
\hline
nic vicorob & 95.1 & 9.18 & 0.137 & 91.0 & 7.65 & 0.37 & 88.5 & 7.15 & 0.43 \\
 \hline
\end{tabular}
\caption{Official iSeg test set results of the top ranking teams. The best values for each metric have been highlighted in bold. Our exclusive multi-label method outperformed the first and second ranked teams~\cite{bui2017dense,dolz2018hyper} at the time of the challenge and stands, overall, in the first place among all first and second round submissions through December 2018 (\href{http://iseg2017.web.unc.edu/results}{iSeg first round}, \href{http://iseg2017.web.unc.edu/evaluation-on-the-second-round-submission}{iSeg second round}). Note that the HD metric is not considered a reliable performance metric for medical image segmentation~\cite{taha2015metrics} as it is very susceptible to outliers.}
\label{table:res}
\end{table*}

\section{Conclusion}
We introduced a new deep densely connected network~\cite{huang2017dense} based on \cite{bui2017dense,jegou2017tiramisu}, called 3D FC-DenseNet that has more depth, more skip connections and less parameters than its predecessors, while having the capability of including 8 times the regular patch sizes ($128 \times 128 \times 128$ vs $64 \times 64 \times 64$) due to its early downsampling and late upsampling layers. To train this deep network we used similarity $F_\beta$ loss functions that generalized the Dice similarity, and achieved better precision-recall trade-off and thus improved performance in segmentation. We designed two pipelines for training, a single-label (regular network without exclusive multi-label) and an exclusive multi-label procedure. Experimental results in 6-month old infant brain MRI segmentation show that performance evaluation metrics (on the validation set) improved by using exclusive multi-label rather than single-label training. The loss function was designed to weigh recall higher than precision (at $\beta=1.5$) for CSF, while using equal precision-recall ratio ($\beta=1$) for WM labels against GM, based on class prevalence in the training set. Official test results based on DSC and ASD scores on the iSeg challenge data show that our method generated the best results in isointense infant brain MRI segmentation, improving the results of all previous DenseNet-based methods~\cite{bui2017dense,dolz2018hyper}.

\section{Acknowledgements}
\noindent This study was supported in part by a Technological Innovations in Neuroscience Award from the McKnight Foundation and National Institutes of Health grants R01 EB018988 and R01 NS079788.

\bibliography{main}

\newpage

\appendix

\section{Hyper-parameter selection}
\label{sec:hyper}
The values of $\beta$ for each class are selected automatically in training based on the ratio of the number of instances per every other class over the number of instances for all classes being equal to the coefficient of all false negatives in equation~(\ref{eq:fbeta}):
\begin{equation}
    \frac{\beta_z^2}{1 + {\beta_z^2}} = \frac{\Sigma_{k=1}^nN_k - N_z}{\Sigma_{k=1}^nN_k} + \lambda \; \Longrightarrow \; \beta_z = \sqrt{\frac{(1+\lambda)\Sigma_{k=1}^nN_k - N_z}{N_z - \lambda\Sigma_{k=1}^nN_k}}
\label{eq:betaselection1}
\end{equation}
\noindent which we saw fit regarding the necessary sensitivity rate for each class based on the complement of its portion on all classes. $\beta_z$ denotes the chosen value for the $\beta$ hyper-parameter for label z, $N_z$ corresponds to the total number of labels for class z, $n$ is the number of classes and $\lambda$ is an extra recall hyper-parameter which we set to 0.1 for this experiment. If we assume $\lambda$ of 0, then equation~(\ref{eq:betaselection1}) becomes the square root of the reverse ratio between the target label and all other labels:
\begin{equation}
\beta_z\;\bigg|_{\lambda=0}= \sqrt{\frac{\Sigma_{k=1}^nN_k - N_z}{N_z}}
\label{eq:betaselection2}
\end{equation}
\noindent Based on prevalence rates of $21.84\%$ for CSF and $31.45\%$ for WM, and $\lambda = 0.1$, $\beta$ values of 1.5 and 1 were approximated and used in this study for CSF and WM classes, respectively.

\section{Evaluation metrics}
\label{sec:metrics}
To evaluate and compare the performance of the network against state-of-the-art in isointense infant brain MRI segmentation, we report the Dice Similarity Coefficient (DSC):
\begin{equation}
\mathrm{DSC} = \frac{2\left | P\cap R \right |}{\left | P \right |+\left | R \right |} = \frac{2TP}{2TP+FP+FN}
\end{equation}

\noindent which is equivalent to the $F_1$ score calculated by setting $\beta=1$ in Equation~(\ref{eq:fbeta}). $TP$, $FP$, and $FN$ are the true positive, false positive, and false negative rates, respectively; and $P$ and $R$ are the predicted and ground truth labels,
respectively.
In the iSeg challenge, in addition to the DSC score, Hausdorff Distance (HD) and Average Surface Distance (ASD):
\begin{equation}
\mathrm{HD} = \max\{\underset{q \in R}{\max} \; d(q, P), \underset{q \in P}{\max} \; d(q, R)\}
\;\;,\;\;
\mathrm{ASD} = \frac{1}{|R|+|P|}\Big(\underset{q \in R}{\sum} d(q, P) + \underset{p \in P}{\sum} d(p, R)\Big)
\end{equation}
were also reported in the test set results, where $d(q, P)$ denotes the point-to-set distance: $ d(q, P) = \underset{p \in P}{\min} ||q - p ||$, with $||.||$ presenting the Euclidean distance and $|.|$ denoting the cardinality of a set. Average Surface Distance (ASD), also known as Mean Surface Distance (MSD) is the average of all the distances from points on the boundary of $P$ to the boundary of $R$ and vice versa, while HD only accounts for the maximum distances between predictions and ground truths. According to \cite{taha2015metrics}, HD is generally sensitive to outliers and because noise and outliers are common in medical segmentations, it is not recommended to use HD directly. For example, broken lines that frequently occur in HD calculation on complex shapes increase the HD measure and cause mismatches. Consequently, the use of HD is highly discouraged and is not an appropriate and reliable unit of measure in medical image segmentation where it involves point-to-set matching on complex shapes, which is a procedure that is prone to errors and is susceptible to outliers.


\section{Tables and Figures}
\label{sec:tablesfigures}

\setcounter{figure}{0}
\renewcommand{\thefigure}{S\arabic{figure}}

\begin{figure}[h]
\begin{subfigure}
    \centering
    \includegraphics[width=0.45\textwidth]{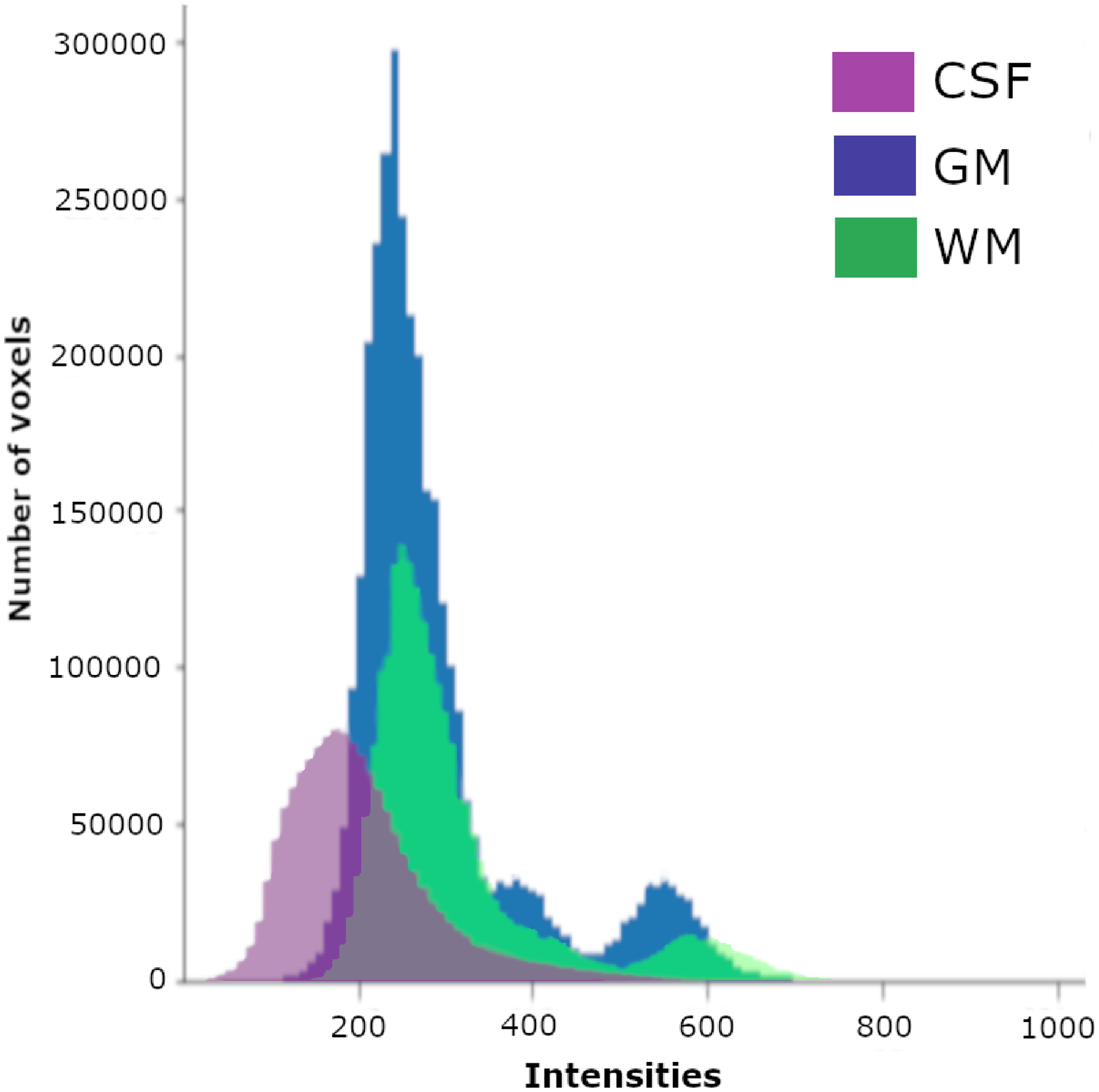}
\end{subfigure}%
\begin{subfigure}
 \centering
    \includegraphics[width=0.45\textwidth]{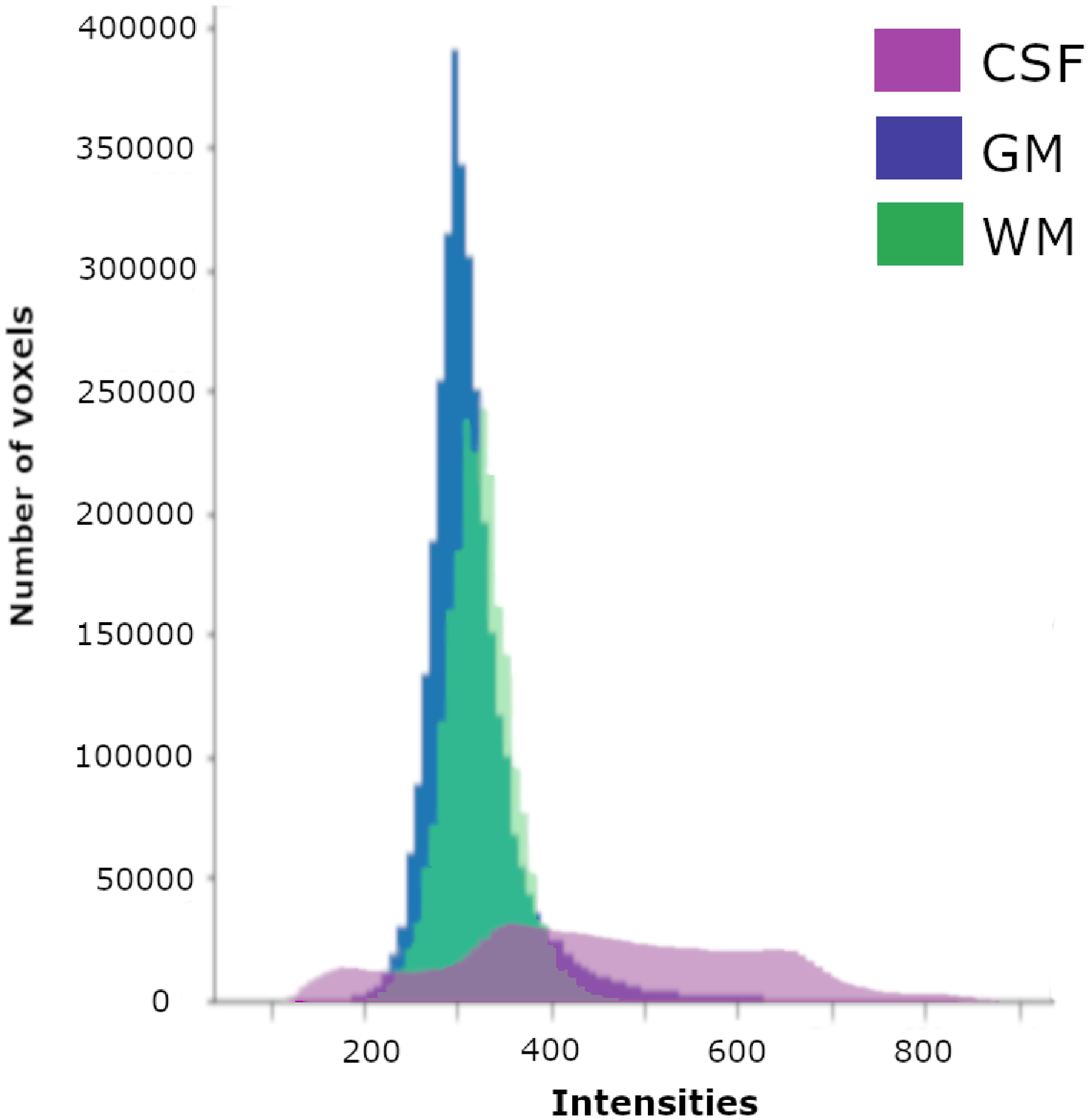}
\end{subfigure}
\caption{Intensity distributions of all three classes on a) T1-weighted and b) T2-weighted MRI scans on all images in the iSeg training set. At 6 months of age, the intensity values of the white matter (WM) and gray matter (GM) in the infant brain, considered in iSeg, largely overlap. This makes WM-GM segmentation on these images very challenging. Given both scans the cerebrospinal fluid (CSF) shows much better separation from WM and GM based on image intensity values.}
\label{fig:intensity}
\end{figure}

\setcounter{table}{0}
\renewcommand{\thetable}{S\arabic{table}}

\begin{table*}[h!]
\small
\centering
{\setlength{\tabcolsep}{0.9em}
 \begin{tabular}{|c||c|c|c|c|c|} 
 \hline
 \multirow{2}{*}{Method} & \multicolumn{3}{c|}{DSC} & \multirow{2}{*}{Depth} & \multirow{2}{*}{Params}\\ \cline{2-4}
& CSF & GM & WM &&\\
\hline
3D Unet \cite{cciccek20163d} & 94.44 & 90.73 & 89.57 & 18 & 19M\\
\hline
DenseVoxNet \cite{Yu2017densevox} & 93.71 & 88.51 & 85.46 & 32 & 4.34M \\
\hline
DenseSeg - MSK SKKU \cite{bui2017dense} & \textbf{94.69} & \textbf{91.57} & \textbf{91.25} & 47 & 1.55M \\
\hline
\hline
\hline
3D FC-DenseNet (Single-label) & 94.86 & 91.18 & 89.27 & 60 & 1.5M\\
\hline
3D FC-DenseNet Exclusive Multi-label & \textbf{95.19} & \textbf{91.79} & \textbf{90.37} & 60 & 1.4M \\
 \hline
\end{tabular}}
\caption{Average performance metrics (on validation sets) of several state-of-the-art methods trained and evaluated on the iSeg challenge dataset. The best values for each metric have been highlighted in bold. The top three methods in the table are derived from~\cite{bui2017dense} with training process and cross validation folds that are different from our methods in the bottom two rows, so the top and bottom parts of the table cannot be directly compared. Comparable results on the official challenge test dataset are shown in Table~\ref{table:res}. This table shows relative performance of DenseNets, 3D Unet, and DenseVoxNet style network architectures and their depth and number of parameters. Paired sample t-test between our exclusive multi-label and single-label trained models (bottom two rows) confirmed that differences were statistically significant ($p < 0.05$).}
\label{table:comp}
\end{table*}

\end{document}